# Automated Essay Scoring Using Transformer Models

Sabrina Ludwig [1,*], Christian Mayer [1], Christopher Hansen[2], Kerstin Eilers[3], and Steffen Brandt [4]

[1]   University of Mannheim; sabrina.ludwig@uni-mannheim.de; mayer@bwl.uni-mannheim.de
[2]   Kiel, Germany; info@christopher-hansen.de
[3]   Kiel, Germany; kerstin_eilers@gmx.de
[4]   opencampus.sh; steffen@opencampus.sh
[*]   Correspondence: sabrina.ludwig@uni-mannheim.de

**Abstract:**

Automated essay scoring (AES) is gaining increasing attention in the education sector as it significantly reduces the burden of manual scoring and allows ad hoc feedback for learners. Natural language processing based on machine learning has been shown to be particularly suitable for text classification and AES. While many machine-learning approaches for AES still rely on a bag-of-words (BOW) approach, we consider a transformer-based approach in this paper, compare its performance to a logistic regression model based on the BOW approach and discuss their differences. The analysis is based on 2,088 email responses to a problem-solving task, that were manually labeled in terms of politeness. Both transformer models considered in that analysis outperformed without any hyper-parameter tuning the regression-based model. We argue that for AES tasks such as politeness classification, the transformer-based approach has significant advantages, while a BOW approach suffers from not taking word order into account and reducing the words to their stem. Further, we show how such models can help increase the accuracy of human raters, and we provide a detailed instruction on how to implement transformer-based models for one's own purposes.

## Contents

## 1. Introduction

Recent developments in Natural Language Processing (NLP) and the progress in Machine Learning (ML) algorithms have opened the door to new approaches within the educational sector in general and the measurement of student performance in particular. Intelligent tutoring systems, plagiarism-detecting software, or helpful chatbots are just a few examples of how ML is currently used to support learners and teachers [1]. An important part of providing personalized feedback and supporting students is automated essay scoring (AES), in which algorithms are implemented to classify long text answers in accordance with classifications by human raters [2]. In this paper, we implement AES using current state-of-the-art language models based on neural networks with a transformer architecture [3,4]. In doing so, we want to explore the following two main questions:

(1) To what extent does a transformer-based NLP model produce benefits compared to a traditional regression-based approach for AES?
(2) In which ways can transformer-based AES be used to increase the accuracy of scores by human raters?

In a first step, we provide information on current work in the field of AES and the different methods of conducting automated essay scoring. Since transformer models are a very promising development, we further provide a more detailed explanation on their basic characteristics and their differences from the approaches used previously. By providing the most relevant technical details for the implementation, we also aim to provide practical guidance for the use of transformer-based models for one's own purposes. We then present the results of an empirical example based on student responses to a problem-solving task submitted as an email in a simulated office environment. Finally, we will conclude with a discussion of the formulated questions, the future relevance of transformer-based models in AES and their limitations.

## 2. Methodological Background for Automated Essay Scoring and NLP based on Neural Networks

### 2.1 Traditional Approaches

Automated essay scoring has a long history. In the early 1960s, the Project Essay Grade system (PEG), as one of the first automated essay scoring systems was developed by Page [5–7]. In the first attempts, four raters defined criteria (proxy variables) while assessing 276 essays in English by 8th to 12th graders. The system uses a standard multiple regression analysis involving the defined proxy variables (text features such as document length, grammar, or punctuation) representing the independent variables with the human-rated essay score being the dependent variable [8]. With a multiple correlation coefficient of .71 as an overall accuracy for both the training and test set and given the good prediction of the human score for the test set based on derived weights from the training set, the computer reached comparable results to humans [7]. However, with a correlation of .51 for the average document length [7,8] the PEG could be easily manipulated by writing longer texts [2,9]. Thus, this kind of score prediction considers only surface aspects and ignores semantics and the content of the essays [8].

In subsequent studies, the n-gram model based on the bag-of-words approach (BOW) was commonly used for a number of decades (e.g.[10–13]). BOW models extract features from student essays, in the case of the n-gram model, by counting the occurrences of terms consisting of n words. They then consider the number of shared terms between essays of the same class and model their relationship [10,14]. An often-cited AES system using BOW is the Electronic Essay Rater (e-rater for short) developed by [15]. The e-rater predicts scores for essay responses and was originally applied to responses written by non-native English speakers taking the Graduate Management Admission Test (GMAT; 13 considered questions) and the Test of English as a Foreign Language (TOEFL; two considered questions). Two features used in the e-rater are extracted by content vector analysis programs named EssayContent and ArgContent. While the former uses a BOW approach on the full response, the latter (ArgContent) uses a weighted BOW approach for each required argument in a response. Using these features alone results in average accuracies of .69 and .82, respectively. Including all 57 predictive features of the e-rater, the accuracy ranges from .87 to .94 (the number of responses varies between 260 and 915 for each question with a mean of 638; a Kappa value is not reported).

The above models are based on laborious, theory-based feature extraction methods. Therefore, methods such as Latent Semantic Analysis (LSA) were established [16]. LSA are corpora-specific and trained on texts that are related to the given essay topic. They provide a semantic representation of an essay, which is then compared to the semantic representation of other similarly scored responses. In this way a feature vector, or embedding vector, for an essay is constructed, which is then used to predict the score. Foltz, Laham, and Landauer [17] used LSA in the Intelligent Essay Assessor system and compared the accuracy of the system against two human raters. The analysis is based on 1,205 essay answers on 12 diverse topics given in the GMAT. The system achieves a correlation of .70 with the human rating, while the two human ratings show a correlation of .71 (a Kappa value was not reported).

## 2.2 Approaches Based on Neural Networks

The current possibilities for automated essay scoring are largely influenced by the advances in NLP based on neural networks. In this section, we therefore first want to provide a short overview of the dynamic development in this area in the past few years. Since the terminology used in the machine learning context is often different from that used in psychometrics, and is also sometimes ambiguous, beforehand some initial corresponding comments on the meaning of some of the terms used and some additional methodological background will be provided. Thereafter, more detailed background on two important NLP approaches based on neural networks are given along with their corresponding achievements in AES.

### 2.2.1 Terms and General Methodological Background

*Model*

Model is an ambiguous term in the context of machine learning. On the one hand, it can merely refer to a particular architecture for a machine learning model; on the other hand, it can also refer to a model architecture that has already been pre-trained with a specific training dataset, that is, that contains the model weights from the pre-training. Therefore, unless it is clear from the context, we will use the terms architecture or checkpoint[1], respectively, instead of the term model.

Unless it is clear due to the context, we will therefore use the terms architecture, or checkpoint, respectively, instead of the term model.

*Token/ Tokenization[2]*

The initial input for an NLP task is usually a text sequence. The first step is to turn this text sequence into a sequence of numbers, which is then interpretable for the neural networks. This is done by splitting the text into different tokens, assigning each token a number, and storing the token-number relationship in a dictionary. The corresponding process is called tokenization and can be word-based, character-based, or subword-based. Word-based tokenization is typically the starting point for the traditional approaches described above. To reduce the huge amount of possibly occurring words in a sentence, it is thereby usually necessary to reduce the amount of variation in words by removing stop words or using a technique called stemming, which ignores conjugations, declinations, or plurals and only considers the stem of a word (see, for example, [20]). In a character-based tokenization the text is split into characters rather than words. This approach has the advantage that the dictionary of possible tokens is usually very small, and that there are far fewer out of vocabulary tokens in comparison to the word-based approach, where the dictionary must be restricted to a certain maximum of words and where all other words will be tokenized as unknown or simply removed. The disadvantage, however, is that each token only incorporates a very limited amount of information (for Chinese languages, for example, this is different though), and the tokenization of an input sequence will lead to a large amount of input tokens for the model with a corresponding computational burden. Subword tokenization is a combination of these two approaches, in which words (sometimes even word pairs or groups) that occur very frequently are not split and other words are split into smaller subwords. For example, by splitting the word "clearly" into "clear" and "ly" or the word "doing" into "do" and "ing". There are different subword segmentation algorithms available for such tokenization. Popular ones are WordPiece [21], which is, for example, used in the BERT model [4], which we will apply for the essay scoring task demonstrated

---

in this paper; Byte-Pair-Encoding (BPE) [22] as used in the GPT-3 model [23]; or SentencePiece [24] or a unigram-based approach [25], which are used in many multilingual models.

*Padding*

Neural networks exclusively work with input sequences of equal length. After tokenizing the input sequence, the second important step is therefore the padding, which yields an equal length of all input sequences. The length of the input sequence is given by the architecture of the used neural network that is used. The padding step then either truncates each input sequence to a predefined length (if it is longer than the predefined length) or completes it with special empty tokens (if it is shorter than the predefined length).

*Embeddings*

While the tokenization turns the text sequence into a sequence of numbers, the embeddings provide the tokens with a meaning by turning the single number into a high dimensional vector (in the case of the GPT-3, for example, the embedding vector has a dimensionality of 12,288 [23]). The embedding vectors can be calibrated either in a supervised way, by comparing each token's meaning in relationship to a result for a particular NLP task, or in an unsupervised way, by using token-to-token co-occurrence statistics (like, for example, the GloVe embeddings [26] or by using a neural net with an unsupervised training task. A defining characteristic of current language models (see section below) and of their embeddings is how this unsupervised training task is defined. In general, though, one or several tokens are hidden, and the model—including the embeddings—is then trained (i.e., calibrated) to predict the hidden token. Besides the embeddings for the meaning of a single tokens, transformer models additionally include embeddings for the position of a token within an input sequence, and sometimes also for segments of tokens, where the embeddings provide information on the probability that a certain series of tokens is followed by another series of tokens, or (in the case of a supervised training) that a series of tokens has an equivalent meaning to another series of tokens [4,23].

*Feature*

A feature is an independent variable in an ML model. In an essay scoring task, the features typically consist of an embedding for each token of a text sequence.

*Label*

The dependent variable in an ML model is called a "label" while in an essay scoring task, this is usually called the "score".

*Language Model*

In NLP, a language model describes a model trained to predict the next word, subword, or character in a text sequence. The best language models are currently all based on transformer architectures [27]. More details on transformers are given in the corresponding chapter below.

### 2.2.2 Methodological Background on Recurrent Neural Nets (RNN)[3]

RNNs are a relatively old technique in terms of neural network structures. The two most prominent types of RNNs are the Gated Recurrent Units (GRU) [28] and the Long Short-term Memory (LSTM) [29]. The corresponding models are largely responsible for breakthrough results in speech recognition and were also dominated the field of NLP until the appearance of transformer-based models. While the GRU is very efficient for smaller datasets and shorter text sequences, the LSTM also allows longer text sequences to be modelled  by integrating a memory cell that can pass information relatively easily from one point to another within a sequence. In particular the LSTM thereby allowed the training of language models with contextual embeddings, and is embeddings where the position of a token in a sequence would make a difference for its meaning  [30]. Due to their sequential structure RNNs in general though have the disadvantage that training them is very slow. Furthermore, both GRU and LSTM suffer from the "vanishing gradient problem", which describes an optimization problem that traditionally occurs in large neural nets and thereby makes the training of LSTMs for very long sequences, for example, impractical [31].

### 2.2.3 Results of Recurrent Neural Networks for Automated Essay Scoring Tasks

---

[3]    For a more in-depth introduction, we recommend Andrew Ng's course on RNNs: https://www.coursera.org/lecture/nlp-sequence-models/recurrent-neural-network-model-ftkzt

Table 1 provides exemplary studies examining similar tasks. The results for RNNs using the dataset published in 2012 on Kaggle for the AES competition under the name "Automated Student Assessment Prize" (ASAP). Taghipour and Ng [32] applied RNNs based on self-trained word embeddings, and their combination of a CNN and a bidirectional LSTM model significantly outperforms the two baseline models, based on support vector regression and Bayesian linear ridge regression, while also outperforming models based on just the LSTM, the CNN, or a GRU.

Alikaniotis and colleagues [33] also analyzed the abovementioned Kaggle dataset and propose a new, score specific embedding approach in combination with a bidirectional LSTM yielding a quadratic weighted Kappa of .96. However, Mayfield and Black [34] are critical of this approach stating that it leads to skewed statistics by grouping questions with maximum score ranging from 4 point to 60 points into a single dataset.

**Table 1.** Exemplary Studies on Recurrent Neural Networks for Automated Essay Scoring Tasks

| Study | Task | Data | Model | Kappa |
|-------|------|------|-------|-------|
| Taghipour & Ng (2016) | Scoring essay answers to 8 different questions, some of which depend upon source information | 12,978 essays with a length of 150 to 550 words | LSTM + CNN | .76 |
| Alikaniotis et al. (2016) | Scoring essay answers to 8 different questions, some of which depend upon source information | 12,978 essays with a length of 150 to 550 words | LSTM combined with score-specific word embeddings | .96 |

### 2.2.4 Methodological Background on Transformer Models

For the application of a transformer model, it is fundamental to understand the three basic architecture types of a transformer: encoder models, decoder models, and encoder-decoder models. Table 2 provides a brief overview of the three types, the tasks they are predominantly used for, and exemplary models that have been implemented and trained based on the respective type.

In general, transformer models are neural networks based on the so-called attention mechanism and were originally introduced in the context of language translation. The attention mechanism was presented in 2014 by Bahdanau et al. [35]. They showed that instead of encoding a text from the source language into a vector representation and then decoding the vector representation into the text of the target language, the attention mechanism allows to avoid this bottleneck of a vector representation to be avoided between the encoder and decoder by allowing the model to directly search for relevant tokens in the source text when predicting the next token for the target text.[4]

In 2017, Vaswani et al. [3] then showed that "Attention is All You Need". While the encoder and decoder models for translation tasks before were mainly based on RNNs, the authors demonstrated that not only can the described vector representation be replaced by an attention mechanism but the encoder and decoder models themselves can be implemented based on the attention mechanism alone. They implemented a self-attention mechanism, in which different attention layers tell the model to focus on informative words and neglect irrelevant words. They showed that this way the model achieves new performance records on several translation benchmarks while having a fraction of the training cost compared to the best models previously used.[5]

---

[4]    For a detailed introduction to the attention mechanism, we recommend Andrew Ng's lecture on Attention Model Intuition: https://www.youtube.com/watch?v=SysgYptB198

A major advantage of the transformer architecture is that it allows parallel processing of the input data and is not affected by the vanishing gradient problem. This makes it possible to train with larger and larger datasets, resulting in better and better language models [36]. Today, transformer-based architectures are applied to a large variety of tasks not only in NLP but also in fields like computer vision, predictions based on time series or tabular data, or multi-modal tasks [37–40]. And in the majority of fields transformer-based models are currently leading the benchmarks [42]. The additional capabilities of the model come with a price, however. While models pretrained with enormous datasets increase the performance, the computational processing of billions of parameters is costly and time-consuming [43]. Moreover, it restricts the training of such models to large corporates or governmental institutions and could therefore prevent independent researchers or smaller organizations from gaining equivalent access to such models. [44].

**Table 2.** Transformer Architectures

| Architecture | Examples | Tasks |
|---|---|---|
| Encoder | ALBERT, BERT, DistilBERT, ELECTRA, RoBERTa | Sentence classification, named entity recognition, extractive question answering |
| Decoder | CTRL, GPT, GPT-2, GPT-3, Transformer XL, GPT-J-6B, Codex | Text generation |
| Encoder- decoder | BART, T5, Marian, mBART | Summarization, translation, generative question answering |

*Note.* Adapted from the "Hugging Face Course", summary of chapter 1 (https://huggingface.co/course/chapter1/9).

The encoder-decoder architecture describes a sequence-to-sequence model as proposed in the original "Attention is All You Need" paper [3]. This type of architecture is particularly trained to transform (in the case of an NLP task) a text sequence of a certain format into a text sequence of another format, such as in translation tasks.

The encoder architecture includes (as indicated by the name) only the input, or left-hand side, of the original transformer architecture and transforms each input sequence into a numerical representation. Typically, encoder models like BERT, RoBERTa, or ALBERT are specially used for text classification or extractive question answering [4,45,46]. Since AES is a special case of text classification, this type of models is therefore particularly well suited to AES, and accordingly the use of the BERT model has become increasingly prominent in the literature on AES [47].

Finally, the decoder architecture includes only the output, or right-hand side, of the original transformer model. In the original model, the decoder takes the information of the input sequence from the encoder and generates in a stepwise procedure one token after the other for the output. At each step, the decoder not only considers the information from the decoder, but also combines it with the information given by the output tokens that were already generated. Using only the decoder architecture therefore results in a model that generates new tokens based on the tokens it already generated or based on a sequence of initial tokens it was provided with (typically called "prompt"). Decoder models, like the GPT-3 or the Transformer XL, are therefore applied to generate text outputs. By providing specifically designed prompts such models can be used for other tasks such as translation or classification tasks as well though [48].

All mentioned models from the different architecture types have in common that they are, in a first step, trained self-supervised on large text corpora to calibrate powerful language models. In a second step, they are then fine-tuned to a specific task using supervised learning and hand-coded labels.[6] This way the language models can transfer the learned knowledge to any more specific task (transfer learning) [49].[7]

---

[5] For a more thourough understanding, we recommend "The Illustrated Transformer" by Jay Alammar (https://jalammar.github.io/illustrated-transformer/) and the video "Transformer Neural Networks – Explained" (https://www.youtube.com/watch?v=TQQlZhbC5ps).
For an intuitive comparison of the attention mechanism and RNNs, we further recommend: https://towardsdatascience.com/the-fall-of-rnn-lstm-2d1594c74ce0

Thus, innovative approaches like high-dimensionality attention-based transformers moved into the focus as they process sequential data in parallel and are highly context-sensitive.

### 2.2.5 Results of Transformer Models for Automated Essay Scoring Tasks

Table 3 shows results of current research studies using transformers for AES. Rodriguez and colleagues [50] use the same Kaggle dataset as the studies shown in Table 2 and compare two transformer-based models, BERT [4] and XLNet [51], with a variety of other models. In particular, they show that the transformer-based approaches yield results comparable to that of a model combined of LSTM and CNN (see Table 2 above).

Mayfield and Black [34] also analyzed the same reference dataset but considered only five of the eight questions in their analyses arguing that the remaining questions with 10 or more scoring classes each are not representative of overall performance and can skew the reported results. They applied different transformer-based models with DistilBERT [52] yielding the highest Kappa value. However, for the case of the selected five questions from the dataset the n-gram model based on the BOW approach yields a comparable result, while the transformer-based approach results in a large increase in needed computational power.

**Table 3.** Recent Studies on Transformer Models for Automated Essay Scoring Tasks

| Study | Task | Data | Model | Kappa |
|-------|------|------|-------|-------|
| Rodriguez et al. (2019) | Scoring essay answers to 8 different questions, some of which depend upon source information | 12,978 essays with a length of 150 to 550 words | BERT XLNet | .75 .75 |
| Mayfield, & Black (2020) | Scoring essay answers to 5 different questions, some of which depend upon source information | 1,800 essays for each of the five questions, each with a length of 150 to 350 words | N-Gram DistilBERT | .76 .75 |

### 3. Data

To investigate the potential of transformer-based models for AES, we used a dataset from the study "Domain-Specific Problem-Solving Competence for Industrial Clerks ("DomPL-IK") conducted in 2014 with 780 vocational and educational training students [53]. The trainees, enrolled in their second or third year of a three-year commercial apprenticeship program learning, were asked to solve three complex domain-specific problems within the business domain "controlling" and to communicate their decision via email.

In scenario 1, students had to conduct a deviation analysis to compute target costs and the variances between target and actual costs within a spreadsheet. Then they were asked to integrate the computed quantitative results with qualitative characteristics provided within information-enriched documents. Finally, the participant needed to communicate possible consequences to a fictitious colleague via email [53]. Similarly, in scenario 2, participants selected a supplier based on a benefit analysis considering quantitative and qualitative characteristics. And in scenario 3, students needed to decide between in-house production and an external supplier. Distributed over these three business scenarios, the 780 apprentices submitted 2088 non-empty email answers in German, including on average 61.9 words (*SD* = 41.2). The maximum length was 512 words. However, most essays had a length of between 32 and 84 words (*Mdn* = 54, see Figure 1).

---

[6]    An exception are models like the GPT-3, which have been successfully applied to a variety of tasks without further fine-tuning but by designing appropriate prompts only.

[7]    For an introduction on transformers and transfer learning we recommend: https://huggingface.co/course/chapter1/4

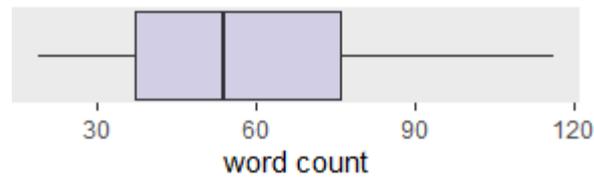

**Figure 1.** Boxplot representing the typical number of words per essay (outliers adjusted).

Furthermore, as in many classification problems, the dataset is very unbalanced: The majority class "polite" includes 92.5% of the cases (n = 1,942) and the minority class "impolite" only 7.5% (n = 146), which will be important to take care of during model training and evaluation.

Considering the human rating, the submitted student emails have been manually scored by trained experts on previously defined scoring rubrics. For the given analyses, we selected an item which classifies whether the response email addressed to the supervisor was written in a polite form, or not. In the study's competence model this item is part of the dimension measuring "Communicating Decisions Appropriately". We assume that this classification is very much task- and domain-independent, and it should therefore be possible to transfer and adapt a corresponding AES to new, previously unseen texts. Emails that contain disrespectful, sarcastic, or rude phrases [54] are scored, or labeled (the more commonly used expression in ML), as "impolite", whereas emails with a courteous tone throughout are labeled "polite". For further details regarding the scoring process please refer to Seifried and colleagues [55] and Brandt, Rausch, and Kögler [56].

## 4. Method

In the following we describe in detail how the regression-based and the transformer-based classification approaches are implemented. The code is implemented in Python since the main packages to train machine learning models are provided in Python, and guidelines and code examples for the latest models are typically also only provided in Python. All Python code used for the analyses presented in this paper is provided in a GitHub repository, which is linked in the appendix.

For readers more familiar with the R infrastructure, we additionally included R markdown scripts in the repository, which allow the Python code to be run using the common RStudio environment.

### 4.1 Basic Data Preparation Used in Both Approaches

The basic dataset includes the two columns "text" and "label", where the column text includes the emails as plain text (no html) with line feeds included as "\n"; the column label includes the integers "0" or "1" representing "impolite" or "polite". The data import depends on the local or cloud infrastructure that is used for the computation and information of the different forms are given in the GitHub repository. The first important step after importing the data is to randomly split it into training (70%) and test (30%) data. A validation dataset is neither in the regression-based nor in the transformer-based approach model needed since we do not perform any hyper-parameter tuning of the models.

For the implementation of the split, we use the function `train_test_split()` from the Scikit-learn package [57]. We use a seed for the random state of the function to assure that the random split stays the same in any repetition.

```
from sklearn.model_selection import train_test_split
train_text_series, test_text_series, train_label_series,
test_label_series = train_test_split(data["text"], data["label"],
test_size = 0.30, random_state = 42)
```

The train_test_split() function returns a Pandas Series object. Since the tokenizer, which will be applied in the next step expects a list object, we convert the data type correspondingly by using the function to_list ():

```
train_text = train_text_series.to_list()
test_text = test_text_series.to_list()
train_label = train_label_series.to_list()
test_label = test_label_series.to_list()
```

### 4.2 Regression Model Estimation

For any machine learning task, it is good common practice to at first consider the results of a baseline model and then consider the benefits of applying a more sophisticated model. Regression analysis is a standard approach for supervised classification tasks [58] and frequently used in the field of AES [59,60].[8]

Before conducting the actual regression analysis, it is necessary to first tokenize the text data given in the responses and then extract meaningful features from them. The tokenization is done using a BOW approach and includes the following two steps:

   (1) removing stopwords, by using the stopword list for German from the NLTK library [61], and

   (2) stemming of all remaining words, by using the Porter Stemmer from the NLTK library [61].

Both steps are implemented in the function `process_mail()` included in the file `utils.py`, which is part of the linked GitHub repository. To extract the features for each response also two steps are conducted:

   (1) creating a frequency dictionary, with the information on how often a word was used in a polite or impolite response, and

   (2) computing for each response a sum score for politeness and for impoliteness, based on the words included in the responses and their values in the frequency dictionary.

For the training data, the corresponding code looks as follows:

```
# Create frequency dictionary
freqs = build_freqs(train_text, train_label)

# Extract features
train_features = np.zeros((len(train_text), 2))
for i in range(len(train_text)):
    train_features[i, :]= extract_features(train_text[i], freqs)
```

With the aforementioned `process_mail()` function included in the `build_freqs()` function. After this, the data frame `train_features` includes the respective sum scores for each of the responses. In the same way the test data is preprocessed.

In the next step, we estimate the logistic regression model by using the function `LogisticRegresion()` from the Scikit-learn package [57].

```
from sklearn.linear_model import LogisticRegression
log_model = LogisticRegression(class_weight= 'balanced').fit(train_features,
    train_label)
```

Classification algorithms work best with balanced data sets. That is, when the number of cases in each class is equal [62]. For unbalanced datasets as the one considered in this paper, it is crucial to take care of this already during training, to not get trivial results. The `LogisticRegression()` function therefore allows to specify an argument `weight`, which yields that each response gets a weight according to the class it belongs to, where responses from smaller classes get higher weights and responses from larger classes smaller ones.

Finally, we evaluate the results of the logistics regression by computing the confusion matrix, the accuracy, the F1 score [63], the ROC AUC [64], and Cohen's Kappa [65] for the predictions from the test data. While accuracy and Cohen's Kappa are common measures in the social sciences to consider the extent of agreement between two ratings, the F1 score and the ROC AUC score are common measures in machine learning. By computing all measures, we want to consider their differences and communalities. The following Python code, yields the computation and export of the measures:

```
from sklearn import metrics

print("Confusion Matrix:\n", metrics.confusion_matrix(test_label,
    log_model.predict(test_features)))
print("Mean Accuracy:\n", log_model.score(test_features, test_label))
print("F1 Score:\n", metrics.f1_score(test_label, log_model.predict(test_features)))
print("ROC AUC:\n", metrics.roc_auc_score(test_label,
    log_model.predict(test_features)))
print("Cohen's Kappa:\n", metrics.cohen_kappa_score(test_label,
    log_model.predict(test_features)))
```

---

[8] Regression Analysis can also be considered as a simple ML approach, as explained by Andrew Ng in his famous "Machine Learning" course: https://www.youtube.com/watch?v=kHwlB_j7Hkc&list=PLLssT5z_DsK-h9vYZkQkYNWcItqhlRJLN&index=4

### 4.3 Transformer-Based Classification

Just as in the regression-based approach, the first step for the classification based on a transformer model is to tokenize the text data. The form of the tokenization is strictly bound to the pretrained transformer that will be used. We use two different pretrained models from the Hugging Face library [66] for our AES task and correspondingly use two different tokenizer functions provided together with these models via the Hugging Face library. The two models, or checkpoints, we use are "bert-base-german-cased" and "deepset/gbert-base", which were released in June 2019 and October 2020, respectively [67]. A main difference between the two models is the size of the dataset used for the pretraining. While the first model is trained on about 12 GB of text data, the newer one is trained on 163 GB of text.

The Python code to tokenize text for the checkpoint "deepset/gbert-base" is as follows:

```
from transformers import AutoTokenizer

checkpoint = "deepset/gbert-base"
tokenizer = AutoTokenizer.from_pretrained(checkpoint)

train_encodings = dict(tokenizer(train_text, padding=True, truncation=True,
    return_tensors='np'))
```

The class `AutoTokenizer` from the `transformer` package includes a function `from_pretrained()`, which allows to instantiate a tokenizer for the selected checkpoint. This tokenizer is then applied on the text data (here for the training data). It splits all texts into subwords according to the WordPiece tokenization method [21], pads or truncates the resulting vectors to the length of the chosen model's context window (i.e., the maximum number of tokens), and assigns an integer ID to each text token. The resulting data is then returned in form of a dictionary. Depending on the model, it might also include some additional advanced methods, which we do not need in this case though but only the plain dictionary. Therefore, we additionally apply the function `dict()` to receive a standard Python dictionary.

Tokenizing text for the checkpoint "bert-base-german-cased" is done by simply changing the text string providing the name of the checkpoint correspondingly.

As mentioned in the previous section on the regression analysis, it is crucial to consider the unbalanced number of elements in the classes. For the fine-tuning of the transformer model, we must explicitly compute and define weights for the two classes.[9] In the two lines of code below we, at first, compute how often each label is present in the training dataset and, then, define a dictionary, which assigns a weight to each label in that way that it compensated for the differences in their frequencies.

```
unique, counts = numpy.unique(train_label, return_counts=True)
class_weight = {0: counts[1]/counts[0], 1: 1.0}
```

---

[9]    In general, there are various other ways to tackle the issue of inbalances data (see, for example, He and Garcia [68]). However, for comparibility reasons, we chose to use the same approach here as in the regression analysis.

The next step is to define the model that will be fine-tuned. We conduct the fine-tuning based on the Keras framework [69]. The definition of the model includes setting various hyperparameters, which are then typically optimized via numerous optimization runs and by evaluating the results with on a validation dataset. However, for our analyses we refrain from optimizing the model's hyperparameters and use the standard setting recommended in the Hugging Face course [19]. Overall, we define the batch size, the number of epochs, the learning rate, the optimizer, the model architecture, and the loss function, and these definitions are then compiled into a single model. As described in the commented code below, the learning rate is defined via a scheduler that is adjusted after each training step, where the change in the learning rate depends on the given batch size and the number of epochs as well as on the size of the training dataset. To obtain the definition of the model architecture and the pretrained weights, we use the function `from_pretrained()` from the class `TFAutoModelForSequenceClassification` of the transformers library. By choosing this class, the Hugging face library provides a TensorFlow (i.e., Keras) version of the provided checkpoint and adds a classification head to it. The argument `num_labels=2` further indicates that the classification head is to be defined for two classes. The code for the model definition then looks as follows (the initial import statements have been omitted here):

```
# Definition of batch size and number of epochs
batch_size = 8
num_epochs = 3

# Definition of the learning rate scheduler
# The number of training steps is the number of samples in the dataset, divided by
    the batch size then multiplied by the total number of epochs
num_train_steps = (len(train_label) // batch_size) * num_epochs
lr_scheduler = PolynomialDecay(initial_learning_rate=5e-5, end_learning_rate=0.,
    decay_steps=num_train_steps)

# Definition of the optimizer using the learning rate scheduler
opt = Adam(learning_rate=lr_scheduler)

# Definition of the model architecture and initial weights
model = TFAutoModelForSequenceClassification.from_pretrained( checkpoint,
    num_labels=2)

# Definition of the loss function
loss = SparseCategoricalCrossentropy(from_logits=True)

# Definition of the full model for training (or fine-tuning)
model.compile(optimizer=opt, loss=loss, metrics=['accuracy'])
```

After the model definition, the fine-tuning of the model is straightforward and conducted with the defined number of epochs, batch size, and weights:

```
model.fit(train_encodings, np.array(train_label),
    class_weight=class_weight, batch_size=batch_size,
    epochs=num_epochs)
```

After fine-tuning the model with the training data, we can then use the model to predict the labels for the test data. This is done in two steps. In the first, we compute for all responses the probabilities of each label, and in the second, we select for each response the label with the highest probability as prediction. To get the probabilities in the first step, we apply a softmax function to the predicted logits[10] that are returned as result from the fine-tuned model. The softmax function normalizes the logits and makes them interpretable as probabilities.

---

[10]    In machine learning the term logits usually refers to the raw predictions that a classification model generates and get passed to the softmax function. It is not referring to the inverse of the sigmoid function here.

```
import tensorflow as tf
test_pred_prob = tf.nn.softmax(model.predict(dict(test_encodings))['logits'])
test_pred_class = np.argmax(test_pred_prob, axis=1)
```

Finally, we compute the confusion matrix and the same four measures of agreement for the predicted test labels, which is equivalent to the computation shown in the regression analysis above.

## 5. Results

The most complete comparison of the agreements of the three considered models with the human ratings provide the confusion matrices shown in Table 4.

**Table 4.** Confusion Matrices

| Actual \ Predicted | Impolite | Polite |
|---|---|---|
| Regression results for test data | | |
| Impolite | 31 | 15 |
| Polite | 87 | 494 |
| German BERT (Jun. 2019) results for test data | | |
| Impolite | 29 | 17 |
| Polite | 35 | 546 |
| German BERT (Oct. 2020) results for test data | | |
| Impolite | 32 | 14 |
| Polite | 24 | 557 |

However, for easier comparison, it is common to compute single scores for each model. Common scores are accuracy, F1, ROC AUC, and Cohen's Kappa, the results of which are shown in Table 5.

**Table 5.** Comparison of Model Performances

| Model | Accuracy | F1 Score | ROC AUC Score | Cohen's Kappa |
|---|---|---|---|---|
| Logistic Regression | .84 | .91 | .76 | .30 |
| German BERT (June 2019) | .92 | .95 | .77 | .52 |
| German BERT (Oct. 2020) | .94 | .97 | .82 | .59 |

The accuracy and F1 score show values above .9 in five out of six cases, indicating a good performance on the given AES task across all three models. However, as described above, the classes are largely imbalanced resulting in skewed results for the F1 score as well as for accuracy (see [70]). More informative measures are therefore the ROC AUC score and Cohen's Kappa.[11] For both measures, the transformer-based approaches yield better values than the regression-based approach, and the larger transformer model yielded the best results for all measures. Considering the achieved values, Cohen suggests values between .4 and .6 as moderate agreement, values between .6 and .8 as substantial agreement, and values between .8 and 1 as almost perfect agreement [72]. The result of the large transformer model is for the given data and according to Cohen's Kappa therefore still slightly below a good performance. Considering the ROC AUC, Mandrekar [73] suggests to consider values between .7 and .8 as acceptable, between .8 and .9 as excellent, and above .0 as outstanding. According to ROC AUC standards the performance of the large transformer model can therefore be considered as good.

---

[11] For more details on the relationship of the ROC AUC and Cohen's Kappa, we recommend [71].

Besides the overall measures on correctly classified responses, the transformer models provide a probability estimate for each prediction. We therefore also analyzed how many of the predictions have estimated probabilities of 95% or more. Of the total 627 responses in the test data, 554 predictions, or 88.3%, have a probability of 95% or more. From these 554 predictions a total of 11 responses are not classified in accordance with the human rating (six are coded as 1 by the machine, but as 0 by the human raters; 5 are coded as 0, but as 1 by the human raters). After carefully reviewing these responses, we found that all these responses were in fact classified incorrectly by the human raters and were not coded in accordance with the coding guidelines.[12]

## 6. Discussion

The results have shown that for the humagiven task of scoring a text according to its politeness, particularly the large transformer-based model reach reasonable results while the regression-based approach struggles to score correctly, particularly for impolite emails. The benefit of applying transformer-based models for AES has recently been investigated by several authors (see section 2.2.5). In this context, Mayfield and Black [34] argue that although the increase in the quadratic weighted kappa is only about 5%, it comes at a large opportunity cost in computational resources. And the question of whether the gain in precision outweighs the computational burden that comes with fine-tuning large transformer models like BERT. At the same time, however, they stress the benefits of transformer models when it comes to analyzing writing style, an area with "uncharted territory for AES". Scoring the politeness of a text, is very much like scoring the style of a text. In the given example with German text, it is crucial, for example, to know if formal or informal speech is used, information that would typically be removed by stemming. Furthermore, it might be important where in the text a certain word is used, for example in the greeting at the beginning or later in a different context. The BOW approach, on which the regression analysis is based on, does not consider word order though. For these reasons, it is not surprising that the transformer-based approaches outperformed the regression-based approach for the given AES task.

The comparison of the two BERT models further shows the importance of the language model that was trained using the BERT architecture. Using a BERT with the currently most up-to-date language model for German increased the accuracy scores significantly in comparison to the one that was state-of-the-art the year before. Kaplan and colleagues [36] have shown that larger transformer models based on more training data will also continue to perform better in the future. It is therefore reasonable to assume that the transformer-based approach will continue to yield better results on AES in the future. The size of the models is increasingly considered a problem though, as mentioned above. Hernandez and Brown [74], however, have shown that the amount of compute necessary for a state-of-the-art neural net in 2012 has decreased by a factor of 44x for a net with the same performance in 2019, which even outperforms the Moore's law rate of improvement in hardware efficiency [75]. This observation is, for example, in line with the work of Sanh and colleagues on DistilBERT, in which they have shown that it is possible to reduce the size of a BERT model by 40%, while retaining 97% of its language understanding capabilities and being 60% faster [52]. We therefore think that the efficiency of the transformer-based models will further increase in the future and additionally that much smaller models with adequate language models for AES will be developed.

As we have seen in the results the scores given by the machine learning algorithms include errors. Introducing a fully automated essay scoring is therefore often viewed with skepticism. However, the first step in getting reliably automated essay scores is having reliable labels for training, that is, reliable scores from human raters. Creating such labels is a very laborious and time-consuming task, and two main reasons are: (1) Often a large part of the rating is relatively easy and clear, but nevertheless they cost time and still allow for typos in the actual coding process; and (2) for borderline cases between two scores, it is very hard to get a clear understanding of where to draw the line. Hence, two raters might have slightly different understandings how to interpret the scoring guidelines or a single rater might deviate in his coding due to the halo effect, where the overall impression of the participant's answers biases the scoring [76]. Furthermore, misclassification can also be caused by reasons such as, a lack of training, unclear coding guidelines, or a lack of motivation. To obtain reliable scores, it is therefore beneficial to have an answer scored not only by one human rater but by two and in case they deviate by maybe even three raters. However, such an approach is very expensive and, in most studies, hardly feasible due to the resulting costs. Training a machine learning model, even a BERT model, is very cheap in comparison. Furthermore, they have the advantage of not only providing a prediction of the score but also a probability estimate of its correctness. For the given example, we saw that the model based on the latest German BERT model is correct on all scores that are estimated with a probability of 95% or more. For 88.3% of the answers this was the case. Hence, with using the machine score for all answers with such high probability estimates, the number of answers that needs to be scored by human raters could be reduced to the, in this case, 11.7% remaining answers. That is, even when assigning each of these answers to three human raters, the costs are still less then when scoring all answers manually by just one person. Moreover, there is a high chance that the scoring will become more accurate, which then in turn leads to better training data and more accurate machine scores.

---

[12]  For data privacy reasons it is unfortunately not possible to share the responses.

## 7. Conclusion

Above, we discussed the benefits of applying a transformer-based model for AES. The most important factor of how much benefit it provides compared to a BOW approach, will always be the nature of the task that is rated. For knowledge tests, it may well be that a BOW approach is sufficient and more efficient to determine whether, for example, all relevant elements are mentioned. In the given case of measuring a communicative skill, we have shown that the difference in performance can be large, though. We think that the same will also hold for tasks, such as reasoning, in which the order of the words is of particular importance. However, we are not aware of any comparable results yet for reasoning.

A further important benefit of the transformer-based approach, or in more general of a language model-based approach, is that we can expect future improvements to AES just by switching from one language model to another. In the same way, we can also easily switch to scoring a task in a different language by changing to a corresponding language model or to a multi-lingual model, which then allows fine-tuning to the same task in different languages simultaneously [77].

Considering the relationship between AES and scoring by human raters, we believe that while replacing human raters with a machine is problematic in many cases, an important role of AES can be to detect possible errors by human raters and automatically score "very easy" texts to free up resources for hard-to-score texts. We have shown that even without further hyper-parameter tuning but just by using the set of recommended standard settings, it is already possible to obtain a model for such purposes. By providing detailed documentation, we hope to motivate others to use such models and improve human ratings.

**Supplementary Materials:** All code used for the analyses shown in the paper and further additional code is available on GitHub under:
https://github.com/LucaOffice/Publications/tree/main/Automatic_Essay_Scoring_Using_Transformer_Models.

**Funding:** This research was funded by the German Federal Ministry of Education and Research, grant number 01DB081119-01DB1123.

**Institutional Review Board Statement:** The study was approved by the Institutional Review Board (or Ethics Committee) of University of Bamberg, Germany.

**Informed Consent Statement:** Informed consent was obtained from all subjects involved in the study.

**Data Availability Statement:** The dataset including the full texts of the response emails is unfortunately not publicly available. Access to a dataset including the scored responses is available on request here: https://www.iqb.hu-berlin.de/fdz/studies/01DB1119_DomPL-IK/?doi=10.5159/IQB_DomPL-IK_v1

**Acknowledgments:** We thank Andreas Rausch and Jürgen Seifried for the extensive discussions we had on the results and for their helpful critical comments on the earlier version of this manuscript. Finally, we appreciate financial support by the German Federal Ministry of Education and Research.

**Conflicts of Interest:** The authors declare no conflict of interest.